\documentclass{article}

\usepackage[english]{babel}

\usepackage[letterpaper,top=2cm,bottom=2cm,left=3cm,right=3cm,marginparwidth=1.75cm]{geometry}

\usepackage[colorlinks=true, allcolors=blue]{hyperref} 

\usepackage{amsmath}
\usepackage{amsfonts}
\usepackage{amssymb}
\usepackage{amsthm}
\usepackage{bm}
\usepackage{bbm}
\usepackage{mathtools}
\usepackage[inline]{enumitem}
\usepackage{mathrsfs} 
\usepackage{wrapfig}
\usepackage{thmtools,thm-restate}
\usepackage{algorithm}
\usepackage{algorithmic}
\usepackage{url}
\usepackage[capitalise,noabbrev]{cleveref}
\usepackage{color}
\usepackage{xcolor}
\usepackage{xspace}
\usepackage[normalem]{ulem}
\usepackage{graphicx}
\usepackage{makecell}
\usepackage{comment}

\theoremstyle{plain}

\def\*{\star}
\DeclareMathSymbol{\mhef}{\mathord}{operators}{`\-}

\usepackage{booktabs}
\usepackage{footmisc}
\usepackage{tablefootnote}

\usepackage{pifont}

\usepackage[round, sort]{natbib}

\usepackage{listings}
\usepackage{xcolor}
\usepackage{mwe}

\definecolor{codegreen}{rgb}{0,0.6,0}
\definecolor{codegray}{rgb}{0.5,0.5,0.5}
\definecolor{codepurple}{rgb}{0.58,0,0.82}
\definecolor{backcolour}{rgb}{0.95,0.95,0.92}

\lstdefinestyle{mystyle}{
    backgroundcolor=\color{backcolour},
    commentstyle=\color{codegreen},
    keywordstyle=\color{magenta},
    numberstyle=\tiny\color{codegray},
    stringstyle=\color{codepurple},
    basicstyle=\ttfamily\footnotesize,
    breakatwhitespace=false,
    breaklines=true,
    captionpos=b,
    keepspaces=true,
    numbers=left,
    numbersep=5pt,
    showspaces=false,
    showstringspaces=false,
    showtabs=false,
    tabsize=2
}

\def\PARADIGM{Learning from Language Feedback\xspace}
\def\paradigm{LLF\xspace}
\def\benchmark{LLF-Bench\xspace}
\def\Benchmark{LLF-Bench\xspace} 

\def\bandit{\texttt{llf-bandit}\xspace}
\def\movie{\texttt{llf-reco-movie}\xspace}
\def\poem{\texttt{llf-poem}\xspace}
\def\optimization{\texttt{llf-optimization}\xspace}
\def\gridworld{\texttt{llf-gridworld}\xspace}
\def\parking{\texttt{llf-parking}\xspace}
\def\metaworld{\texttt{llf-metaworld}\xspace}
\def\alfworld{\texttt{llf-alfworld}\xspace}

\title{\begin{tabular}{cc}\raisebox{-0.4\height}{\includegraphics[height=5\fontcharht\font`A]{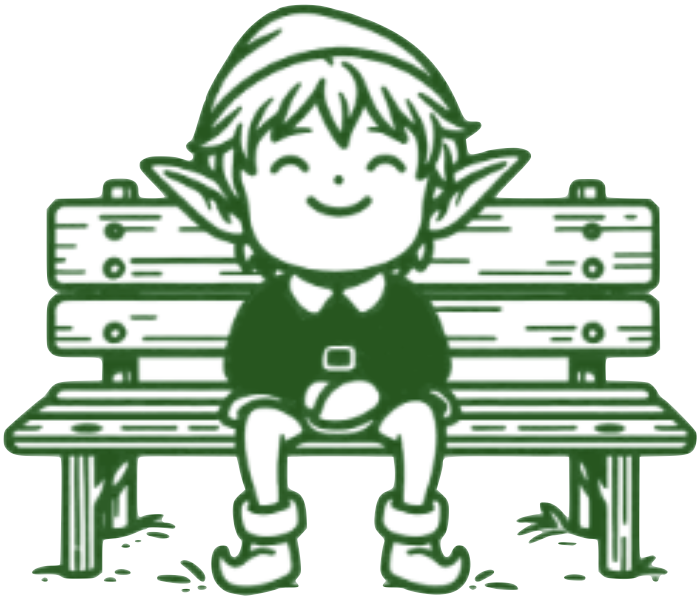}} & \texttt{\Benchmark}: Benchmark for Interactive Learning\\[-0.5cm] & from Language Feedback\end{tabular}}

\author{\hspace*{-0.9cm}
{Ching-An Cheng$^a$\thanks{All authors contributed equally and are alphabetically ordered. Correspondence can be sent to \href{mailto:chinganc@microsoft.com}{chinganc@microsoft.com}, \href{mailto:akolobov@microsoft.com}{akolobov@microsoft.com}, \href{mailto:dimisra@microsoft.com}{dimisra@microsoft.com}, \href{mailto:anie@stanford.edu}{anie@stanford.edu}, \href{mailto:adswamin@microsoft.com}{adswamin@microsoft.com}}}\quad
{Andrey Kolobov$^a$\footnotemark[1]}\quad
{Dipendra Misra$^a$\footnotemark[1]}\quad
{Allen Nie$^b$\footnotemark[1]}\quad
{Adith Swaminathan$^a$\footnotemark[1]}\\[.5cm]
Microsoft Research$^a$\quad  Stanford University$^b$
\\[.5cm]
Website:  \href{https://microsoft.github.io/LLF-Bench}{https://microsoft.github.io/LLF-Bench}
}
\date{}

\begin{document}
\maketitle




\begin{abstract}
We introduce a new benchmark, LLF-Bench (Learning from Language Feedback Benchmark; pronounced as ``\emph{elf-bench}''), to evaluate the ability of AI agents to interactively learn from natural language feedback and instructions. Learning from language feedback (LLF) is essential for people, largely because the rich information this feedback provides can help a learner avoid much of trial and error and thereby speed up the learning process. Large Language Models (LLMs) have recently enabled AI agents to comprehend natural language --- and hence AI agents can potentially benefit from language feedback during learning like humans do. But existing interactive benchmarks do not assess this crucial capability: they either use numeric reward feedback or require no learning at all (only planning or information retrieval). LLF-Bench is designed to fill this omission. LLF-Bench is a diverse collection of sequential decision-making tasks that includes user recommendation, poem writing, navigation, and robot control. The objective of an agent is to interactively solve these tasks based on their natural-language instructions and the feedback received after taking actions. Crucially, to ensure that the agent actually \emph{learns} from the feedback, LLF-Bench implements several randomization techniques (such as paraphrasing and environment randomization) to ensure that the task isn't familiar to the agent and that the agent is robust to various verbalizations. In addition, LLF-Bench provides a unified OpenAI Gym interface for all its tasks and allows the users to easily configure the information the feedback conveys (among suggestion, explanation, and instantaneous performance) to study how agents respond to different types of feedback. Together, these features make LLF-Bench a unique research platform for developing and testing LLF agents.

\end{abstract}

\section{Introduction}


Natural language provides an intuitive medium for a person to teach an AI agent, since that is also how humans learn and teach each other.
Compared with rewards, typically used in the reinforcement learning (RL) paradigm~\citep{sutton2018reinforcement}, language feedback can provide rich signals about the agent's behaviors, in addition to a quantitative measure of instantaneous performance. 
For instance, language feedback can explain why the agent's previous bad behaviors should be avoided, rather than just punishing the agent without giving justification.
Language feedback can also provide direct suggestions on how the agent can improve its future behavior, similar to action feedback used in imitation learning (IL)~\citep{ross2011reduction,spencer2021feedback} but easier to provide --- after all, it's easier said than done. For example, providing action feedback to a robot requires setting up additional teleoperation devices which might not always be feasible, while language feedback can be given verbally by an ordinary user~\citep{liu2023interactive}.

\Benchmark (\PARADIGM Benchmark; pronounced as ``\emph{elf-bench}'') is a simulation benchmark designed to evaluate an AI agent's ability to \emph{learn} interactively from \emph{just} language feedback.
%
\Benchmark is a collection of sequential decision making problems (ranging from item recommendation, poem writing, navigation, to robot control). In each problem, an agent interacts with a task environment and receives language instructions and feedback.
At the start of an episode, the agent is first given a natural language \emph{instruction} that describes the objective of the task, the rules, and (optionally) side information that may help solve the problem. After executing an action in the environment, the agent receives teacher \emph{feedback} in natural language which can be used as a learning signal.

\begin{figure}
    \centering
    \includegraphics[scale=0.2]{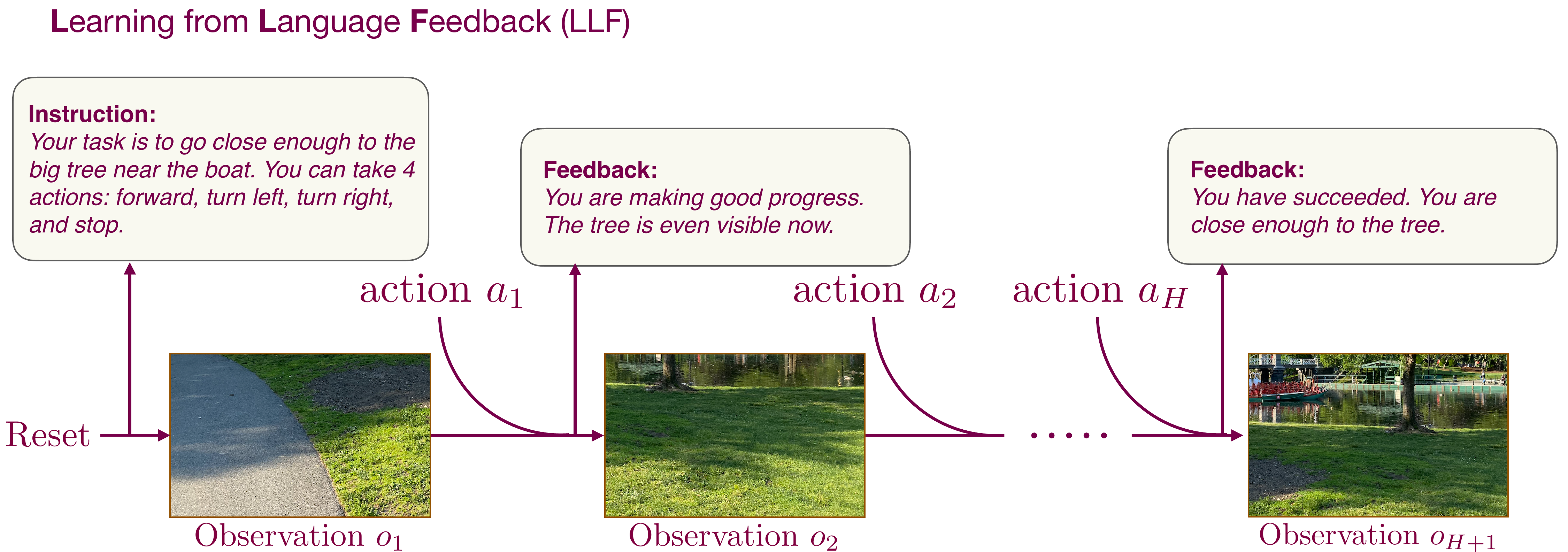}
    \caption{Shows an example navigation task to illustrate our setup, Learning from Language Feedback (LLF). A single episode in LLF starts with a given instruction and can be multi-step long. The actions are taken by the agent that changes the observation and provides a \emph{text feedback} to the agent. The agent receives no reward or any other form of feedback.}
    \label{fig:flowchart}
\end{figure}

\begin{figure}
    \centering
    \includegraphics[width=0.9\linewidth]{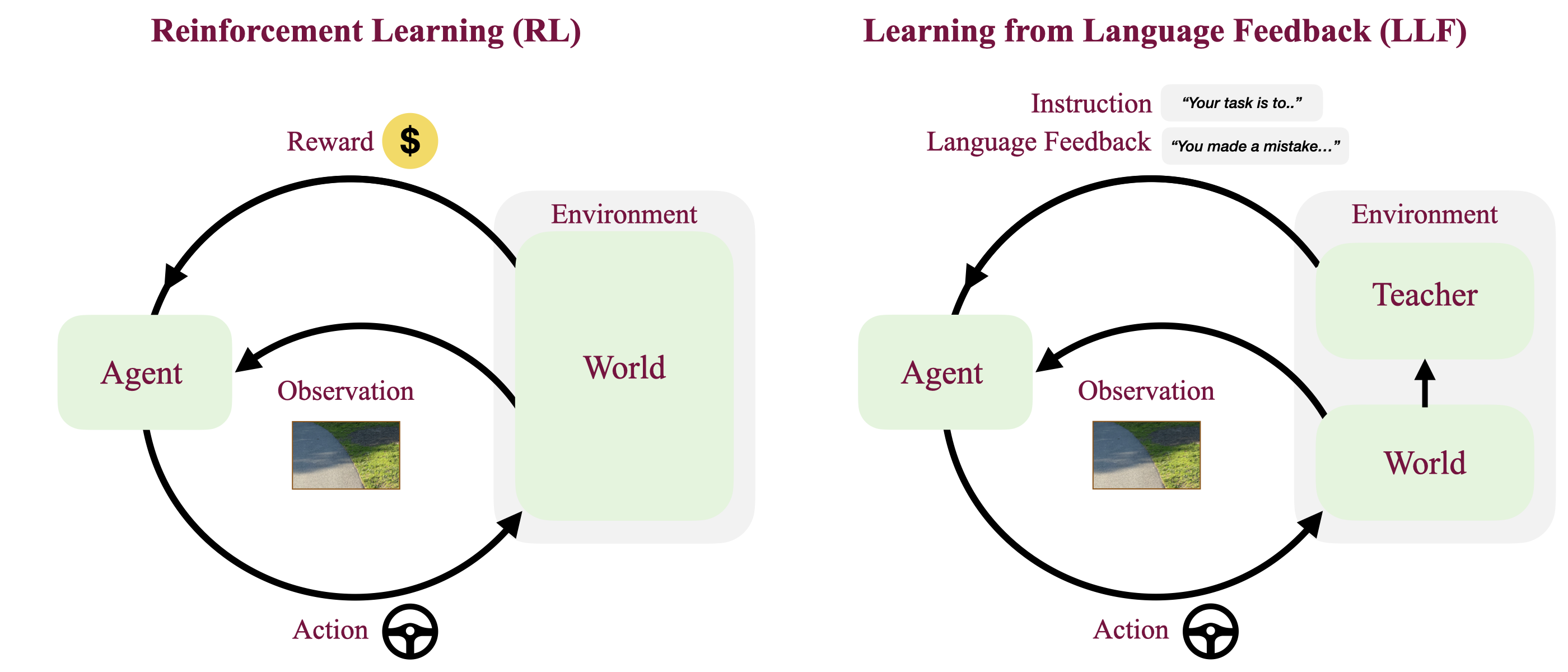}
    \caption{Comparison between RL and LLF setups. LLF replaces reward feedback in RL with language feedback and generalizes the reward maximization objective to general instructions that can be specified via natural language. }
    \label{fig:rl-llf}
\end{figure}

We call this paradigm \emph{\PARADIGM} (\paradigm). \paradigm generalizes reinforcement learning (RL) from reward maximization to general problem solving. Like in RL, \paradigm focuses on sequential decision problems. However, in contrast to RL, an \paradigm agent does \emph{not} receive rewards and is \emph{not} necessarily tasked with maximizing returns. Figure~\ref{fig:flowchart} shows an example \paradigm flowchart. \paradigm replaces RL's assumption of numeric rewards with generic task instructions and feedback expressed in natural language. 
We can recover RL as an instance of \paradigm, e.g., with the instruction \textit{``Maximize the accumulated rewards.''} and the feedback template \textit{``You've received a reward of X.''}.
But \paradigm covers many other scenarios that would be unnecessarily difficult to describe in the conventional RL framing, e.g., training a robotic arm controller by giving it general advice about the types of actions it should consider in certain situations, or asking an agent to write a poem in a certain mood by showing a few examples. We illustrate the similarities and differences between RL and LLF in Figure~\ref{fig:rl-llf}.


\begin{figure}
    \centering
    \includegraphics[width=1\linewidth]{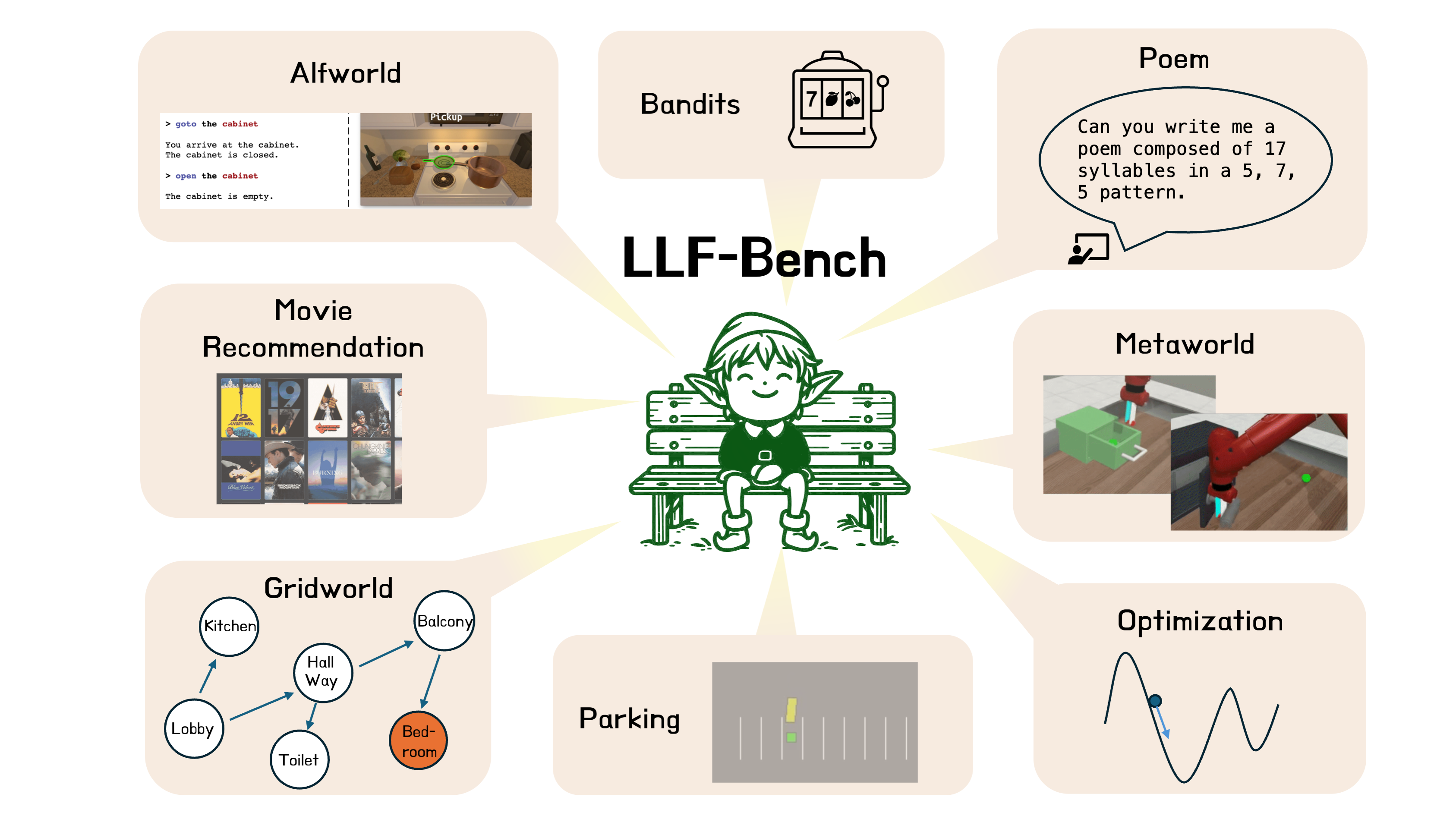}
    \caption[]{\Benchmark (``Elf-bench'') includes 8 sets of \paradigm problems. Image by Bing Chat.}
    \label{fig:enter-label}
\end{figure}

How can an interactive agent use language feedback to achieve efficient learning in the \paradigm setting?
A prerequisite is that the agent can already understand commonsense and reason with natural language, so that the agent can focus on skill learning rather than figuring out the basics of language understanding (because instructions, rewards, and feedback in LLF are written in natural language). This has been the main reason why \paradigm has not received significant attention in the past.
However, recently, Large Language Models (e.g., GPT4~\citep{openai2023gpt4}, Gemini~\citep{gemini2023}) have demonstrated impressive natural language processing abilities. In addition, multiple LLM-agents have shown promising signs of solving text-based problems involving decision making, planning, information retrieval, tool uses~\citep{wang2023voyager,schick2023toolformer,wu2023autogen}.
Therefore, LLMs provide a promising way to work with the general-purpose language feedback in \paradigm. Further, solving \paradigm can also be viewed as a way to measure the ability of LLMs to solve new learning tasks. Notably, \benchmark differs from the vast majority of benchmarks for evaluating LLMs which are either non-interactive, or allow the agent designer to choose how to verbalize the environment, which can lead to prompt hacking (i.e., an LLM-agent overfitted to a specific environment through its prompts).

%




\subsection{Highlights of \Benchmark}

We design \benchmark as a research platform to facilitate the development and testing of \paradigm agents (e.g., LLM-agents). 
\benchmark consists of 8 diverse sets of decision-making problems (see Figure~\ref{fig:enter-label}), with different action spaces (discrete, continuous, and free-form text spaces) and decision horizons:  
\begin{itemize}
    \item \bandit is a verbalized version of the classic multi-armed bandit problem, which we implement based on  \href{https://github.com/JKCooper2/gym-bandits/tree/master}{gym-bandits}. \bandit tests the agent's learning ability in an unknown environment with a finite number of actions.
    \item \poem consists of a set of poem writing tasks, where the agent needs to write a poem satisfying certain syllable- and line-constraints. These problems tests the agent's learning ability to infer and solve constraint satisfaction problems.
    \item \movie simulates a classic recommendation scenario where a user wants movie or TV show recommendations based on some preferences. The user specifies their preferences in text, and any recommendation made by the agent is matched to a movie database for checking whether the preferences are matched correctly.
    \item \optimization consists of 8 loss functions (Rosenbrock, Bohachevsky, etc.) and provides an interface to give verbal feedback for the task of optimization on any loss function.
    \item \parking extends the \href{https://github.com/Farama-Foundation/HighwayEnv}{Highway} gym environment, providing a long horizon goal-conditioned continuous control task. The agent must control an ego-vehicle to park in a given location without colliding with any obstacles in the environment. 
    \item \gridworld evaluates the agent's ability to navigate in a graph-based environment. Each node of the graph is a room and the edges are doors connecting the rooms. The agent's goal is to  
    navigate from the room it starts in to the room with treasure. 
    \item \alfworld adds a wrapper on top of the Alfworld text-based environment~\citep{ALFWorld20} to provide language feedback instead of reward. In $\alfworld$, the agent is tasked to 
    solve 
    problems in a text-based house environment. The agent is tested for generalization as each episode can contain a new task in a new house environment.
    \item \metaworld is a low-dimensional state-based version of the existing Meta-World v2 benchmark~\citep{yu2021metaworld}. It comprises 50 simulated robotic manipulation tasks featuring a Sawyer arm and various objects that this arm needs to bring into desired configurations, such as opening doors, placing cubes in boxes, etc.
\end{itemize}

When designing a \emph{learning} benchmark, an important consideration is whether the evaluation can truthfully reflect an agent's learning and generalization abilities and separate them from overfitting.
To this end, we make two important design choices:
\begin{enumerate}

    \item Following the framing of \paradigm, \benchmark implements the task instruction as part of the environment, as opposed to as part of the agent.
    The latter is common in the current literature of LLM-agents, and many LLM-agents heavily rely on using task-specific prompt templates~\citep{yao2023tree,wang2023voyager}. 
    By this design, we encourage users of \benchmark to develop agents that can simultaneously work well across different problems sets in \benchmark.
    We hope that this paradigm shift would facilitate the development of more generic learning agents that can solve multiple tasks, rather than being engineered for solving just a single task.


    \item \Benchmark provides the option to further randomize the textual description of task instruction and feedback that the agent receives. In addition, for several environments, we randomize the environment's latent parameters (e.g., to permute the action ordering in \bandit or change the room connectivity in \gridworld) when the environment is reset.
    Sensitivity to different phrasing of the same instruction is often used to measure the robustness of a text-based model~\citep{ribeiro2018semantically,wallace2019universal}.
    %
    This design is motivated by the observation that LLMs \emph{as of now}, do not always perfectly understand semantics and can be sensitive to the exact texts that are presented~\citep{zhu2023promptbench}. It has been shown LLMs suffer from recency bias and can give drastic different outputs for semantically similar inputs~\citep{arora2023ask, leidinger2023language}.
    To combat that, for each problem instance in \Benchmark, we manually curate a set of syntax templates via paraphrasing, which are used to produce a diverse yet semantically equivalent set of task instructions and feedback during interactions.
    By introducing randomization, \benchmark can better evaluate the agent's ability of task solving and prevent the agent from overfitting a specific text realization.
\end{enumerate}


One prominent feature of \benchmark is its configurable feedback system.
Taking inspiration from the education research literature~\citep{shute2008focus}, we classify the language feedback into 3 different types:
\begin{enumerate*}[label=\textit{\arabic*)}]
    \item feedback of the current performance (similar to reward scalars and success booleans),
    \item suggestions of future behaviors (e.g., hints or things to avoid)
    \item explanation of past behavior (e.g., why some behaviors are bad and should not be repeated).
\end{enumerate*}
By default, a testing environment in \benchmark provides a mix of these feedback (when appropriate). It can also be easily configured to provide feedback based on any subset of these categories.

For ease of use, \benchmark adopts the OpenAI Gym API~\citep{1606.01540}, which abstracts the interaction with \texttt{reset} and \texttt{step} API functions.
\benchmark environments return the natural language instruction and feedback as the observation (a python dict) and the action spaces vary across problems. \benchmark environments also return rewards per the Gym \texttt{step} API. While agents in the \paradigm setup do not use rewards, the returned rewards can be used to evaluate an \paradigm agent's performance; this feature makes the \benchmark environments also usable as typical RL environments.
\Benchmark also provides a text-mode option (where both the observation and the action are free-form texts), so that it can also be used as a benchmark for evaluating LLM-agents as well.

\subsection{Related Setups and Benchmarks} \label{sec:related work intro}

Many RL environments incorporate natural language. We provide a list summarizing their main features in Table~\ref{tab:related}. The RL environments can use language to describe the reward/goal~(\textbf{instructions}), the \textbf{observations}, or the \textbf{actions}. Commonly, language is used as goal-specifying \textbf{instructions} (which is essentially a reward function) for an RL agent (e.g., GridLU by~\citet{bahdanau2018learning}, ViZDoom Text by~\citet{chaplot2018gated}, ISI Block by~\citet{misra2017mapping}, and Puddle World by~\citet{janner2018representation}). In this context, understanding and mapping instructions/goals to the state of the environment is the key challenge.
Some RL environments naturally have \textbf{observations} in text; these include text-based adventure games~(Text World by~\citet{cote2019textworld} and NetHack by~\citet{kuttler2020nethack}) and HTML webpages~(MiniWoB by~\citet{shi2017world}, MiniWOB++ by~\citet{liu2018reinforcement}, and WebShop by~\citet{yao2022webshop}).
Other RL environments have \textbf{action} spaces in text, i.e. an RL agent can generate a sequence of tokens as an action, such as a structured text representing a short executable program~(e.g. SHRDLURN by~\citet{wang2016learning}). However, this was considered challenging due to the relatively large vocabulary space and the difficulty of learning to generate a sequence.
None of these environments provide rewards as text and do not provide feedback on actions. They also do not consider variations in language expressions -- such as different phrasing or writing that represent the same underlying goal or state of the environment.
Many of these environments are unsuitable for testing LLM agents due to having an observation space that is pixel or vector-based, and the types of tasks are dissimilar to what people use LLMs for today.


On the other hand, several benchmarks have been proposed to evaluate LLM-based agents for decision-making~(AgentBench by~\citet{liu2023agentbench}, OpenAGI by~\citet{ge2023openagi}, MINT by~\citet{wang2023mint}, and LMRL Gym by~\citet{abdulhai2023lmrl}). However, many tasks in these benchmarks center around planning and information retrieval problems. Few require the agent to learn and adapt beyond what an LLM can already do. A real-life user would leverage an LLM-based agent to solve challenging tasks but also give intermediate feedback, such as ``make the title text larger'' or ``wrap the code with an error-catching block.'' \benchmark supports such intermediate feedback as well. Also, due to the lack of language variations, developers might identify a specific prompt that solves an instantiation of the task, over-fitting to a particular writing of the task specification. Lastly, because an LLM-based agent interacts with human users, the specification of reward from a user can often be text. Are LLM-based agents capable of learning and adaptation from rewards represented as text? \benchmark aims to provide a set of environments to help answer this question while addressing the challenges in reliably benchmarking LLM-based agents.






\begin{table}[t]
\centering
\small
\resizebox{\textwidth}{!}{
\begin{tabular}{@{}r|ccccc@{}}
\toprule
          Environment   & \begin{tabular}[c]{@{}c@{}}Observation\\ Space\end{tabular} & \begin{tabular}[c]{@{}c@{}}Action\\ Space\end{tabular} & \begin{tabular}[c]{@{}c@{}}Reward\\ Space\end{tabular} & \begin{tabular}[c]{@{}c@{}}Language\\ Variations\\ (Robustness)\end{tabular} & \begin{tabular}[c]{@{}c@{}}Language\\ Feedback\end{tabular} \\ \midrule
             & \multicolumn{5}{c}{Language Grounding Envs}      \\ \midrule
SHRDLURN~\citep{wang2016learning}     & Vector                                                      & Text                                                   & Scalar                                                 & None                                                                         & No                                                          \\
GridLU~\citep{bahdanau2018learning}       & Pixel                                                       & Discrete                                               & Scalar                                                 & None                                                                         & No                                                          \\
VizDoom Text~\citep{chaplot2018gated} & Pixel                                                       & Discrete                                               & Scalar                                                 & None                                                                         & No                                                          \\
ISI Block~\citep{misra2017mapping}    & Pixel                                                       & Discrete                                               & Scalar                                                 & None                                                                         & No                                                          \\
Puddle World~\citep{janner2018representation} & Pixel                                                       & Discrete                                               & Scalar                                                 & None                                                                         & No                                                          \\ \midrule
             & \multicolumn{5}{c}{Text-based Games}  \\ \midrule
BabyAI~(\textcolor{blue}{C-Boisvert et al., 2018})       & Pixel                                                       & Discrete                                               & Scalar                                                 & None                                                                         & No                                                          \\
Zork~\citep{narasimhan2015language}         & Text                                                        & Text                                                   & Scalar                                                 & None                                                                         & No                                                          \\
TextWorld~\citep{cote2019textworld}    & Text                                                        & Text                                                   & Scalar                                                 & None                                                                         & No                                                          \\
NetHack~\citep{kuttler2020nethack}      & Pixel                                                       & Discrete                                               & Scalar                                                 & None                                                                         & No                                                          \\ \midrule
             & \multicolumn{5}{c}{Web-Navigation Envs}  \\ \midrule
MiniWoB~\citep{shi2017world}      & Pixel/Text                                                  & Disc/Cont                                                    & Scalar                                                 & None                                                                         & No                                                          \\
MiniWOB++~\citep{liu2018reinforcement}    & Pixel/Text                                                  & Disc/Cont                                                    & Scalar                                                 & Observation                                                                  & No                                                          \\
WebShop~\citep{yao2022webshop}      & Pixel/Text                                                  & Text                                                   & Scalar                                                 & None                                                                         & No                                                          \\ \midrule
             & \multicolumn{5}{c}{LLM Agent Benchmark Envs} \\ \midrule
AgentBench~\citep{liu2023agentbench}   & Text                                                        & Text                                                   & Scalar                                                 & None                                                                         & No                                                          \\
OpenAGI~\citep{ge2023openagi}      & Text                                                        & Text                                                   & Scalar                                                 & None                                                                         & No                                                          \\
MINT~\citep{wang2023mint}         & Text                                                        & Text                                                   & Scalar                                                 & None                                                                         & Yes (LLM)                                                   \\
LMRL Gym~\citep{abdulhai2023lmrl} & Text & Text & Scalar & None & No \\ \midrule
\textbf{\benchmark} (Ours)          & Text                                                        & All                                                   & Scalar\tablefootnote{The scalar reward is for evaluation, not for agent learning in the \paradigm setup.}+Text                                            & All                                                                          & Yes (Synthetic)                                             \\ \bottomrule
\end{tabular}}
\caption{We list several decision-making environments that involve natural language. Language is used to instruct model behavior, represent observation, or is part of the action output. ``Language Variations'' refers to whether there are multiple descriptions of the same instruction, observation, or reward. ``Disc/Cont'' means the output is a mix of discrete and continuous variables. \textbf{\benchmark} offers text representation for instruction, observation, and reward, generates paraphrasing to prevent prompt hacking, and offers procedurally generated synthetic feedback for fast and cheap evaluation.}
\label{tab:related}
\end{table}

\section{\paradigm: Learning from Language Feedback}

\paradigm is a learning paradigm that generalizes the RL setup. As shown in \cref{{fig:rl-llf}}, \paradigm is an abstract setup\footnote{Here we follow the convention of \cite{sutton2018reinforcement} that ``anything that cannot be changed arbitrarily by the agent is considered to be outside of it and thus part of its environment.''. Therefore, we consider the (physical or digital) world that the agent has effects on as well as the teacher who provides instructions and feedback to the agent as part of the environment. As per \cite{sutton2018reinforcement}, we note that ``the boundary between agent and environment is typically not the same as the physical boundary of a robot’s or animal’s body. Usually, the boundary is drawn closer to the agent than that.''} that models the interaction between an \emph{agent} (e.g., a learning algorithm), a \emph{world} (e.g., a robot hardware, or a recommendation system based on a database), and a \emph{teacher} (e.g., a person who uses or teaches the agent mentioned in \cref{sec:related work intro}).

\subsection{Setup}
The agent in the \paradigm is prompted by the teacher to complete a task in the world with some natural language \emph{instruction}. The instruction may be different from reward maximization and could include information about how to interpret observations, what valid actions are, and side information (such as examples) that may help the agent solve the problem.
After receiving the instruction, the agent sees the initial observation of the world state, and the agent starts to interact with the world by taking actions within the problem's prescribed action space, which like that in RL can be e.g. a finite space, a continuous vector space, or a free-form text space.
After an action is executed, the world's internal state may change and the agent sees the next observation of the world.
As the agent interacts and learns to solve the task, the teacher would provide natural language \emph{feedback} to guide the agent to learn better based on how the agent performs.  This language feedback is a strict generalization of the reward signal in RL and can provide richer information to help agent learn (e.g., suggestions, explanations, etc.). If we group the world and the teacher in \paradigm together as an abstract \emph{environment}, we see that \paradigm mainly replaces the reward maximization objective and feedback in RL with a generic task instruction and language feedback.
In \benchmark, we simulate \paradigm problems through the OpenAI Gym interface, which we will describe in \cref{sec:gym}.
.

\subsection{Motivation}
The \paradigm setup is motivated by the inefficiency and unnaturalness of communicating intentions with rewards.
%
The concept of reward maximization in RL, while giving a simple abstraction of interactive learning, often creates a barrier for humans to transfer knowledge and convey their intention to AI agents.
Reward feedback compresses all the information that one wishes to convey down to a single numerical value, representing signals for encouraging or penalizing certain behaviors.
In addition, rewards are received only after the agent takes actions, so the agent would have to not only learn the task solving skill but \emph{also} learn to understand the task's objective.
%
This bottleneck limits the information that can be transferred to the agent and couples skill learning and intention understanding, causing the agent to learn inefficiently in a trial-and-error manner.

Moreover, in many cases, it is difficult for human designers to fully understand the long-term effects of reward maximization, even when each instantaneous reward makes sense. This misalignment has led to many surprising behaviors of RL agents~\citep{amodei2016concrete}.
Consequently, reward engineering has been a common practice in building RL systems, where the user iteratively tweaks the reward to give to the agent by observing how the agent behaves after maximizing the current reward function.
However, reward engineering is an expensive process. If agents can learn directly from language feedback (i.e. efficiently solving \paradigm), learning systems can be built more economically.

Overall, compared to RL, \paradigm encapsulates the rich language feedback that is used in human to human learning. The expressive nature of this rich language feedback, provides a potentially more efficient mechanism for training agents than RL.


\subsection{\paradigm and RL with text/language observations}

Interactive learning settings with language-based instructions \citep{misra2018mapping,chen2019touchdown} or observations has been extensively studied \citep{wang2016learning,guu2017language,zhong2021silg} in the literature. However, in all these settings, one assumes access to either gold actions or rewards that are crucially necessary for understanding the textual instructions and observations, and learning the optimal policy. In contrast, in \paradigm the agent is neither provided gold actions nor rewards. This makes \paradigm appear as a harder learning setting than RL.
We argue that this difficulty working with general-purpose language feedback, has been the reason why \paradigm hasn't received much attention previously despite its potential benefits.

%

However, with the recent success LLMs at general-purpose language reasoning, we have a powerful tool to work with language feedback and develop \paradigm agents. As mentioned earlier, language feedback can provide more expressive forms of feedback, thereby, providing a more sample-efficient setup for developing agents. For example, the feedback in \paradigm can be formative which is defined by \citet{shute2008focus} as ``information communicated to the learner that is intended to \emph{modify} the learner’s thinking or behavior for the purpose of \emph{improving} learning.''.

%

We also highlight that with access to accurate LLMs, \paradigm is not harder than a RL setting, if the task instructions in \paradigm are detailed enough such that from the observations alone (without language feedback) it is possible for the LLM to infer if the agent has succeeded at following the instruction. Note that this assumption does not mean that the instruction necessarily shows the agent how to solve the problem. 
Under this assumption, \paradigm problems can always be solved without the feedback, by a reduction to a RL problem with sparse binary reward of success (the binary reward can be computed using a LLM to detect success based on the instruction and the observation). However, such a reduction approach would lead to inefficient learning. The main research question of \paradigm is how to best leverage the language feedback, which can convey more information than just success/failure, to learn the optimal policy for the task in a sample-efficient manner.
We next describe \Benchmark as our proposed research platform to measure progress in answering this question.



%

\section{Gym Interface of \Benchmark} \label{sec:gym}

\Benchmark formalizes a wide variety of decision-making problems by extending the popular OpenAI Gym API. The API contains three key functions --- \verb|make|, \verb|reset|, \verb|step| --- that are semantically similar to their Gym namesakes and detailed below.
\begin{itemize}
\item \verb|make|: Returns an \emph{Environment} object similar to \verb|gym.make|. An \Benchmark \emph{Environment} extends classic Gym Environments (e.g., with well-defined \verb|ActionSpace| and \verb|ObservationSpace|) with two additional concepts, \verb|instruction| and \verb|feedback|, that are explained below.
\item \verb|reset|: After an environment is initialized using \verb|make|, it should be \verb|reset| to receive the initial \emph{Observation} from the \emph{Environment}. \Benchmark \emph{Observation} is a python dictionary containing \verb|gym.Observation| (i.e., an observation that is contained in the \verb|environment.ObservationSpace|) as well as \verb|instruction| and \verb|feedback| keys. If the environment uses randomization, then the random number generator can be seeded with the \verb|seed| parameter as input.
\item \verb|step|: Takes as input an action that is contained in the \verb|environment.ActionSpace|, and returns a \Benchmark \emph{Observation} dictionary which includes the \verb|instruction| and \verb|feedback| keys. In addition to the \emph{Observation}, \verb|step| also returns scalar \emph{reward}, boolean flags \emph{truncated} and \emph{terminated} and a miscellaneous \emph{info} dictionary which have the same semantics as \href{https://gymnasium.farama.org/}{Gymnasium} environments. An agent for \Benchmark is expected to solve tasks using the feedback contained within \emph{Observation}, \textbf{without} using the \emph{reward} signal. Signals like \emph{reward} and \emph{info} are provided for backward compatibility with Gymnasium and for automated evaluation.
\end{itemize}
Note that under the hood, \Benchmark implements all \emph{Environment} objects as compatible with the \href{https://gymnasium.farama.org/}{Gymnasium} standard. We provide \verb|EnvironmentCompatibility| wrappers if the \emph{Environment} is instead otherwise compatible with the deprecated Gym (pre-$0.21$ version) standard. We similarly include \verb|TextWrapper| wrappers that can convert any \Benchmark \emph{Environment} with bespoke \verb|ObservationSpace| and \verb|ActionSpace| into one with text as the observation and action spaces. This wrapper allows one to directly interface LLM-based agents with \Benchmark environments and assess their learning and decision-making behavior.

\begin{lstlisting}[language=Python,caption={Sample python code snippet for interacting with \Benchmark environments.},label=fig:sample_code]
import llfbench as gym

# Environments in the benchmark are registered following
# the naming convention of verbal-*
env = gym.make('verbal-gridworld-v0')

done = False
cumulative_reward = 0.

# First observation is acquired by resetting the environment
observation = env.reset()

while not done:

    # Observation is dict having 'observation', 'instruction', 'feedback'
    # Here we print the observation and ask the user for an action
    action = input( observation['observation'] + '\n' +
                    observation['instruction'] + '\n' +
                    observation['feedback'] + '\n' +
                    'Action: ' )

    # Gridworld has a text action space, so TextWrapper is not needed
    # to parse a valid action from the input string
    observation, reward, terminated, truncated, info = env.step(action)

    # reward is never revealed to the agent; only used for evaluation
    cumulative_reward += reward

    # terminated and truncated follow the same semantics as in Gymnasium
    done = terminated or truncated

print(f'Episode reward: {cumulative_reward}')

\end{lstlisting}


Although each \verb|step| also returns a scalar \emph{reward}, the convention we follow (and recommend to users of \Benchmark) is that the agent never sees the reward. It can only access the information in \emph{observation}, \emph{instruction} and \emph{feedback} to decide its actions (e.g., see line 17 in Listing~\ref{fig:sample_code}).

\subsection{Instruction and Feedback}

\emph{Instruction} is a string that is defined inside the \emph{Environment} and describes in natural language the problem that a decision-maker must solve. We recommend that agent-designers should not inspect and overfit to a specific instruction describing the desired task in an environment; the default behavior of \Benchmark environments is to paraphrase instructions in different ways to minimize the chances of prompt overfitting.
Three different types of \emph{Instruction} are supported in \Benchmark, and can be toggled by passing an appropriate \verb|instruction_type| to the \verb|make| command of a \Benchmark environment:
\begin{itemize}
\item Basic: \verb|instruction_type = `b'|. This is the default instruction type for \Benchmark environments. The instructions provide an agent with the goal, semantics of its action space, as well as the expected syntax of its responses. The instruction provides enough information for a competent agent (e.g., a literate human) to begin interacting with the environment.
\item Complete: \verb|instruction_type = `c'|. The instructions additionally provide information to reliably infer (e.g., by a literate human) an optimal policy for achieving the goal.
\item Practical: \verb|instruction_type = `p'|. It contains the \emph{Basic} instructions, and additionally includes \emph{Feedback} for previously executed actions. The goal of a learning agent is to infer the optimal policy (i.e., comparable in performance to the one with \verb|instruction_type = `c'|) as quickly as possible.
\end{itemize}

\emph{Feedback} is a string that provides the signal for an agent to learn from its interaction. \Benchmark implements two kinds of feedback: an atomic feedback, and a composite feedback. The type of feedback an environment provides to an agent is set by passing an appropriate \verb|feedback_type| parameter to \verb|make|. Atomic feedbacks are inspired by the education research literature~\citep{shute2008focus}. \Benchmark currently supports $5$ different types and we plan to include new styles (to include e.g., questioning) in the future:
\begin{itemize}
    \item \verb|feedback_type = `r'|: This is the textualization of the scalar reward signal or success signal from classical RL. By using the text-wrapper and this feedback type, several classical RL environments (implemented in OpenAI Gym or Gymnasium) can be comparably tested with \paradigm agents in \Benchmark.
    \item \verb|feedback_type = `hp'|: This \emph{hindsight positive} feedback provides an explanation about a past action by the agent that was desirable.
    \item \verb|feedback_type = `hn'|: This \emph{hindsight negative} feedback provides an explanation about a past action by the agent that was undesirable.
    \item \verb|feedback_type = `fp'|: This \emph{future positive} feedback provides a suggestion for a potential future action that could be desirable.
    \item \verb|feedback_type = `fn'|: This \emph{future negative} feedback provides a suggestion for potential future actions that should be avoided.
\end{itemize}

\verb|feedback_type = `r'| corresponds to the \textbf{current performance} evaluation from the education research literature, whereas \verb|feedback_type = `fp', `fn'| correspond to \textbf{future behavior} suggestion. Finally, \verb|feedback_type = `hp', `hn'| correspond to the \textbf{past behavior} explanation style of feedback studied in the education research literature.

Composite feedback types allow the environment to provide the agent multiple kinds of atomic feedbacks. This makes for a more realistic learning problem, rather than the same type of atomic feedback at every \verb|step| of the environment.
\begin{itemize}
\item \verb|feedback_type = `a'|: \emph{All} of the Atomic feedback types that are supported by the environment are provided to the agent at each round of interaction.
\item \verb|feedback_type = `m'|: The agent receives a \emph{Mix} of different atomic feedbacks. A random subset of the supported feedback types are sampled by \Benchmark to provide to the agent at each step.
\item \verb|feedback_type = `n'|: The agent receives \emph{No} feedback, this mode is provided for debugging purposes.
\end{itemize}

The \verb|make| API accepts any of the composite feedback types, or any subset of the atomic feedback types to allow fine-grained control of the learning signal that an agent can receive from \Benchmark environments. The default behavior in \verb|make| for any environment uses \verb|feedback_type = 'a'|.
See Listing~\ref{fig:sample_code} for sample code that creates a \Benchmark environment and instantiates an agent to interact with it using \verb|make|, \verb|reset| and \verb|step| API calls.

\subsection{Instruction and Feedback Randomization}
To reduce the sensitivity of learning agents to a specific text realization, \Benchmark implements a template-based paraphrasing system, by which users can randomize the instruction and the feedback that the agent receives.
For each problem in \Benchmark, we implement about 4-20 paraphrased templates for each instruction and each feedback type. When the randomization options are turned on, the \Benchmark environment will randomly choose one from these curated templates to formulate the language instruction and feedback returned to the agent. \Benchmark also provides the option to deterministically use a particular template. The randomness of paraphrasing can be controlled by setting the \texttt{seed} parameter in the OpenAI Gym \texttt{reset} function.

\section{Tasks in \Benchmark}

\Benchmark consists of 8 different problem sets, ranging from user-recommendation, poem-writing, navigation, to robot control.
In the \paradigm setup, the reward is masked out (though the environments in \Benchmark still return rewards for evaluation purposes).
To solve these problem efficiently, an \paradigm agent needs to have sufficient common sense understanding of the natural language instruction and the feedback.
In addition, the agent needs to be able to \emph{learn} from environmental interactions and feedback. We intentionally design these suites of problems such that, while the agent can tell success from the instruction and the environmental observation, it is difficult for the agent to infer the optimal policy from them without additional learning.

These problem sets feature different action spaces, problem horizons, and test different abilities of \paradigm agents. We provide a summary in \cref{tab:problem_sets} and next
describe each problem set in more detail. 

\begin{table}
    \centering
    \begin{tabular}{l|c|c|c|c|c}
         \textbf{Problem Set} & \textbf{Action Space} & \textbf{Horizon} & \textbf{Stateful} & \textbf{Instruction} & \textbf{Feedback}
         \\
         \hline
         \bandit & Discrete & 1 & No & b, p, c & all  \\
         \poem & Text & 1 & No & b & all\\
         \movie & Text & 1 & No & b & all   \\
         \optimization & Continuous & 10 & Yes & b & all  \\
         \parking & Continuous & 100 & Yes  & b & r, hp, hn\\
         \gridworld &  Finite & 20 & Yes & b, p, c & all   \\
         \alfworld & Text & 100 &  Yes & b & all   \\
         \metaworld & Continuous &  30  &  Yes & b & r, hp, hn, fp \\ \hline
    \end{tabular}
    \caption{Properties of problem sets included in \Benchmark. Instruction and Feedback column denote the types of instruction and feedback that are supported by the environment. If feedback is all, then it means that all 5 feedback (r, hn, hp, fn, and fp) are supported. 
    }
    \label{tab:problem_sets}
\end{table}

\subsection{\bandit}

\bandit is a verbalized version of the classic multi-armed bandit problem. We built \bandit based on \href{https://github.com/JKCooper2/gym-bandits/tree/master}{gym-bandits} by adding natural language task instruction and feedback. There are a total of $8$ bandit problems in \bandit. For each problem, the task instruction tells the task name from the underlying \texttt{gym-bandits}, that the goal is a bandit problem, as well as the feasible actions.
While being a bandit problem, \bandit's feedback does not necessarily convey the reward value in text (it depends on the configuration of the feedback type).
When \texttt{reset}, the environment randomizes the order of actions and, if applicable, the underlying reward function. 
The agent here needs to learn to explore and exploit in multiple rounds of interactions to find the best arm as fast as possible with small regret (measured in terms of the hidden rewards). Overall, \bandit  tests the agent's learning ability in an unknown environment with a finite number of actions.

\subsection{\poem}

\poem is a collection of text-generation tasks requiring a poem to be written that conforms to a particular number of lines and number of syllables for each line. Even though there are many types of formal poems, the current set of tasks supports basic types that follow syllable and line-based constraints. Such formal poems include Haiku (a three-line short poem following a 5-7-5 syllable pattern), Tanka (a five-line short poem following a 5-7-5-7-7 pattern), and custom environments where a user can specify the number of lines and how many syllables per line. We use the CMU Pronouncing Dictionary for syllable verification\footnote{\href{http://www.speech.cs.cmu.edu/cgi-bin/cmudict}{http://www.speech.cs.cmu.edu/cgi-bin/cmudict}}. \poem provides detailed fine-grained feedback on each line -- a good environment to test whether the LLM-based agents can improve quickly given feedback.

\subsection{\movie}

\movie is an environment that simulates user-recommendation system interactions on the topic of recommending movies. To simulate a user, the environment will first randomly sample a user preference profile over a set of attributes such as the type of entertainment (TV show or movie), year range (80s, 90s, 2000s, or recent), preferred genres (Action, Comedy, Documentary, etc.), and age restriction (child/family-friendly or R-rated). An agent needs to recommend a few items (no restriction on the number of items) that all satisfy the stated preference. An item-by-item feedback is provided in this environment to point out detailed preference violations that can allow LLMs to improve their recommendations.

\subsection{\optimization}

\optimization provides an easy-to-use interface with automatic procedurally generated feedback that examines LLMs' ability to make a series of proposals $x$ to minimize a particular loss function $y=f(x)$.  The feedback provided in this environment is created by computing gradient $\frac{d y}{d x}$ and then verbalizing this information based on the change in input between the previously proposed $x$ and the current chosen $x$.  We provide implementations of 8 classic loss functions (Rosenbrock, Bohachevsky, etc.), and the base class is easily extendable to cover other loss functions. This is an environment where we can measure LLM's ability to make decisions with observed information on an unknown loss landscape.

\subsection{\parking}
\parking extends the \href{https://github.com/Farama-Foundation/HighwayEnv}{Highway} gym environment to \Benchmark. It is a long horizon goal-conditioned continuous control task where the agent can manipulate the throttle and steering input to an ego-vehicle. It must park the ego-vehicle in a given location without colliding with any obstacles in the environment. We extended the environment by (1) describing the observation and action spaces in text, and (2) verbalizing the per-time-step reward (distance to goal) to provide text feedback about goal progress and obstacle collisions. An agent must learn how its control inputs affect the vehicle's dynamics, and plan to accomplish the eventual parking goal.

\subsection{\gridworld}

The $\gridworld$ domain models a navigation agent in a graph-based gridworld. The world is represented by a graph where rooms are denoted by nodes and edges denote doors. A room can have at most 4 doors along the north, south, east and west direction. These directions form the agent's action space. At any given time, the agent is in exactly one of the rooms. If the agent takes an action, such as $a=\emph{north}$, then it will transition from its current room, to the room connected by the door along the north direction, if one exists. If no such door exists, then the agent stays in the same room. All transitions are deterministic. A room can contain many different types of objects. A unique room, called the treasure room, contains the treasure object. The agent starts in a start room and its goal is to navigate to the treasure room. The number of rooms, objects, object type, and distance to the treasure, can be easily customized.

The agent's observation describes the current room including all the objects in it, and the different doors that are available.
 The agent can get all 5 types of feedback: \verb|r|, \verb|hn|, \verb|hp|, \verb|fn|, and \verb|fp|. For example, for the \verb|fn| feedback, the agent will be told which action, i.e., a door, to avoid going through in the next step.

\subsection{\alfworld}

The $\alfworld$ environment is a wrapper built on top of the popular \verb|AlfWorld| text-game environment~\citep{ALFWorld20} which itself is built as a parallel to the embodied \verb|Alfred| dataset~\citep{ALFRED20}. $\alfworld$ contains multi-step reasoning tasks, where in each episode, the agent is given an instruction in a house setting and must take a sequence of actions to fulfill this instruction. In each step, the agent is given a textual description of what it sees and a list of valid actions. The agent generates a text action (e.g., \emph{open drawer 1}), which if it is valid can change the agent's observation, and if it is invalid then results in no change. The agent additionally gets a reward for each action.
The goal of the agent is to maximize the total reward by solving the task. Unlike the $\gridworld$ setting, the agent is tested for generalization as each episode can contain a new task in a possibly new house environment.

The main addition in $\alfworld$ is the capability to provide text feedback instead of reward. 
The text feedback is generated using an optimal trajectory for that episode, as well as the instantaneous reward and the list of valid commands for each time step. Similar to $\gridworld$, the agent can get all 5 types of feedback: \verb|r|, \verb|hn|, \verb|hp|, \verb|fn|, and \verb|fp|.

\subsection{\metaworld} \metaworld wraps the low-dimensional fully observable state-based version of the existing Meta-World v2 benchmark~\citep{yu2021metaworld} into a textual interface. Meta-World consists of 50 simulated robotic manipulation tasks, in each of which a robotic Sawyer arm needs to move an object into a specified position, e.g., push a puck to a goal location or press a button. An agent trying to accomplish an \metaworld task is presented with an instruction stating that the task is about getting a robotic manipulator to successfully handle an object and explaining what each dimension of the agent's 4D state space means. By default, the environment interprets an agent's action as a target pose where the arm should move\footnote{The dynamics of \metaworld differs from the one in the original Meta-World. Here the agent controls the target location (the simulator runs the P-controller to act in the original Meta-World environment for several steps until the target location is reached or it is timed out), whereas in the original environment the agent controls force to incrementally change the end-effector.
This design is to make the problem horizon shorter and more closely mimic the common use cases of industrial robotic manipulators.
}, and tries to move the arm there using Meta-World's built-in P-controller. At each time step, the agent receives as observation a description of the current state, mentioning the pose of the arm, and all relevant objects in the scene. The language feedback here may include advice on where to move the arm next and where not to move it.



\section{Related Work}









\paragraph{Grounded Language Learning}
Reinforcement learning with textual information has been studied under the branch of multi-modal representation learning. This branch of study has several focuses that are both similar and different from our goal with \benchmark. One focus deals with ambiguity and difficulty in understanding instructions or goals specified by natural language~\citep{wang2016learning,bahdanau2018learning,chaplot2018gated}. While the ambiguity of instructions is a concern, we focus more on robustly behaving under different instructions that all represent the same underlying goal. Another focus of this body of work is to ground visual information with textual instruction -- a core aim of multi-modal representation learning~\citep{bisk2016natural,misra2017mapping}, with an extension to robotic interaction~\citep{karamcheti2022lila,karamcheti2023language}. Language provides a natural shared representation that enables easier transfer between different tasks~\citep{hanjie2021grounding} or supplies important information such as safety constraints for a policy~\citep{yang2021safe}. In previous work, feedback is often not considered. When feedback is considered, it is usually framed as error messages from a syntax parser (if the action space is text) and can indeed be incorporated into learning~\citep{cote2019textworld}. This type of feedback corresponds to \verb|feedback_type = `hn'| in our setup.

\paragraph{Text-based Games} Extending from using reinforcement learning for solving complex games, there are many text-based games that include challenges such as the navigation of space, manipulation of the environment to achieve goals, and reaction to random events. \cite{narasimhan2015language} repurposed a classic text adventure game, Zork, where both observation and action space are text. \cite{cote2019textworld} proposed a set of text-based game environments and included a few carefully designed challenges for RL to solve, such as large state and action space (determined by the vocabulary size) and long credit assignment. On the other spectrum, \cite{kuttler2020nethack} created a learning environment from the game NetHack. Although the game state is represented with hundreds of text symbols,  policy learning is conducted on the screenshot of the terminal. Similarly, BabyAI~\citep{chevalier2018babyai} is a set of procedurally generated grid-like maze environments -- the objects and representation in the environment are a fixed set of symbols. None of these environments consider providing language feedback on the agent's action.

\paragraph{Learning from Language Feedback}
Providing feedback to an RL agent's action as part of the learning signal beyond task rewards has been studied in robotics.
However, most of the efforts were limited to eliciting binary preference feedback~\citep{sadigh2017active,biyik2018batch} or ranking-based feedback from real people~\citep{basu2019active}.
\cite{sumers2021learning} crowd-sourced a small feedback dataset on a small game. They considered three types of feedback, evaluative feedback (which corresponds to \verb|feedback_type = `r'|), descriptive feedback (which in our setup is decomposed into \verb|feedback_type = `hp', `hn'|), and imperative feedback (which corresponds to \verb|feedback_type = `fp', `fn'|). They then used a sentiment classifier to extract coarse information from this feedback to improve the policy's behavior. \cite{nguyen2021interactive} proposed an approach to map textual instructions to trajectories in embodied settings by assuming that a user can label a generated trajectory with the instruction that is likely to generate the trajectory under the optimal policy. More recently, \cite{cui2023no,liu2023interactive} studied the case of language feedback as corrections to a robotic arm at any time of the task execution, which is an instance of the \paradigm setup that we are considering.


\paragraph{LLM Sensitivity to Prompts} A long line of work has investigated smaller-scale language-based systems' sensitivity to different expressions that have the same underlying meaning. They can be categorized as adversarial attacks to text-based systems~\citep{ribeiro2018semantically,wallace2019universal} or as mechanisms to improve language-based systems' output via self-consistency~\citep{edunov2018understanding}. More recently, the lack of robustness to prompts has been found on large language models as well~\citep{liu2023jailbreaking,wolf2023fundamental}. \cite{zhu2023promptbench} proposed a benchmark dataset to investigate the robustness of LLMs on different types of prompts that can contain user errors for tasks related to natural language.

\paragraph{LLM Agent Benchmarks}
Unlike previous efforts to incorporate language to develop RL policies, building agents using LLM has ushered in a new set of challenges. In general, the environments included in these benchmarks only concern planning and information retrieval with sparse reward signals at the end. Very few of these benchmarks measure the ability of an agent to learn and adapt to a task (e.g., the Abstraction and Reasoning Corpus by~\citet{chollet2019measure}).   \cite{liu2023agentbench} proposed a set of environments that cover a few popular types of task setups, such as web browsing, game, and code generation. Their focus is on the diversity of tasks, not the robustness of LLMs or how well they can incorporate feedback, which is dissimilar to how LLMs are currently being used in a user-centered environment. \cite{ge2023openagi} constructed a set of tasks where LLMs are prompted to use language or vision-related models to solve a complex task that requires multiple steps. The task-level feedback they provide is a numerical score from a domain-specific evaluation method. MINT~\citep{wang2023mint} is a benchmark that also offers natural language style feedback. However, MINT synthesizes user feedback by prompting LLMs. This incurs additional costs, introduces additional variability in the evaluation process, and makes it challenging to represent the diversity of human feedback styles. LMRL Gym~\citep{abdulhai2023lmrl} provides a set of 8 environments that include full and partial observability. The tasks are similar to language-grounding tasks and text games. However, no interim feedback is provided during multi-round interactions.

\subsection*{Acknowledgement}

We are thankful to Marc-Alexandre Cote and Victor Zhong for helping us understand the Alfworld environment. We thank Christine Herlihy for helpful discussions on the Movie Recommendation environment. We gratefully acknowledge Ahmed Awadallah and John Langford for their organizational support. Part of this work was done when Allen Nie was an intern at Microsoft Research. We also appreciate ChatGPT and GPT4 for providing some of the paraphrases used in \Benchmark and BingChat for generating the elf-bench image.

\bibliography{ref}

\end{document}